\documentclass{article}
\usepackage[nonatbib, preprint]{neurips_2023}
\usepackage[numbers]{natbib}

\usepackage[utf8]{inputenc} % allow utf-8 input
\usepackage[T1]{fontenc}    % use 8-bit T1 fonts
\usepackage{hyperref}       % hyperlinks
\usepackage{url}            % simple URL typesetting
\usepackage{booktabs}       % professional-quality tables
\usepackage{amsfonts}       % blackboard math symbols
\usepackage{nicefrac}       % compact symbols for 1/2, etc.
\usepackage{microtype}      % microtypography
\usepackage{xcolor}         % colors
\usepackage{placeins}

\usepackage{amsmath}
\usepackage{amsfonts}
\usepackage{amsthm}
\usepackage{amssymb}
\usepackage{enumitem}
\usepackage{bbm}

\DeclareMathOperator*{\argmin}{arg\,min}

\newcommand{\bb}[1]{\mathbb{#1}}
\renewcommand{\cal}[1]{\mathcal{#1}}
\newcommand{\mb}[1]{\mathbf{#1}}

\newcommand{\F}{\cal{F}}
\newcommand{\f}{\mathbf{f}}

\newcommand{\R}{\bb{R}}

\newcommand{\x}{\mb{x}}

\newcommand{\eps}{\varepsilon}

\newcommand{\lr}[1]{\left(#1\right)}

\usepackage{framed}
\usepackage{xcolor}

\renewenvironment{leftbar}[2][\hsize]
{
    \def\FrameCommand
    {
        {\color{#2}\vrule width 3pt}
        \hspace{0pt}
    }
    \MakeFramed{\hsize#1\advance\hsize-\width\FrameRestore}
}
{\endMakeFramed}

\usepackage{subcaption}
\usepackage[capitalize]{cleveref}
\usepackage{adjustbox}

\title{Measuring Feature Sparsity in Language Models}

\author{Mingyang Deng\thanks{Equal contribution} \\ MIT \\
\And Lucas Tao\footnotemark[1] \\ Stanford University \\
\And Joe Benton\thanks{Corresponding author, \texttt{benton@stats.ox.ac.uk}} \\ University of Oxford
}

\begin{document}

\maketitle

\begin{abstract}
Recent works have proposed that activations in language models can be modelled as sparse linear combinations of vectors corresponding to features of input text. Under this assumption, these works aimed to reconstruct feature directions using sparse coding. We develop metrics to assess the success of these sparse coding techniques and test the validity of the linearity and sparsity assumptions. We show our metrics can predict the level of sparsity on synthetic sparse linear activations, and can distinguish between sparse linear data and several other distributions. We use our metrics to measure levels of sparsity in several language models. We find evidence that language model activations can be accurately modelled by sparse linear combinations of features, significantly more so than control datasets. We also show that model activations appear to be sparsest in the first and final layers.
\end{abstract}

\section{Introduction}

Over the past decade, neural networks have demonstrated remarkable performance in many domains, including natural language processing \citep{brown2020language, openai2023gpt}, computer vision \citep{wang2022yolov7}, protein structure modelling \citep{jumper2021highly} and complex strategy games \citep{silver2017mastering, berner2019dota}. However, our capacity to interpret these models lags far behind. Being able to reliably and automatically determine the features used by such models -- that is, the properties of the model inputs that the model extracts and uses for its predictions -- would be a major step forward in our ability to interpret and safely deploy such models.

Many methods have been proposed for extracting model features, including feature visualization \citep{erhan2009visualizing, olah2017feature}, saliency maps \citep{simonyan2014deep} and feature importance scores \citep{lundberg2017unified}. Feature extraction for language models is less well developed, though recent techniques can locate facts within models \citep{meng2022locating} and show how models perform certain linguistic operations \citep{geiger2021causal}. Nevertheless, most approaches to language model interpretability require human input at key stages and scale poorly to extracting many features.

One obstacle is \textit{polysemantic neurons} in language model MLP layers -- neurons that respond to multiple unrelated features \citep{olah2020zoom, elhage2022toy}. One hypothesised explanation for polysemanticity is that in the absence of co-occuring features, models use one neuron to represent several features without loss of accuracy. Toy models for this phenomenon, known as superposition, have been proposed \citep{elhage2022toy, scherlis2022polysemanticity}. These models make two key assumptions: the \textit{linear representation hypothesis}, that activations  in neural networks can be decomposed as a linear combination of vectors corresponding to individual features, and the \textit{sparsity hypothesis}, that only a small number of features are active in any input \citep{elhage2022toy}.

Under these assumptions, decomposing language model activations into components corresponding to different features is an instance of sparse coding \citep{olshausen1997sparse, elhage2022toy, sharkey2022taking}. Recent works have applied sparse coding to automatically decompose language model activations into sparse linear combinations of interpretable feature vectors \citep{yun2021transformer, cunningham2023sparse}. However, these methods have assessed their decompositions by how easy it is to find plausible descriptions of the set of inputs on which a particular found feature is active. Such metrics can indicate how successful these methods are at finding interpretable feature directions, but say little about the accuracy of the underlying assumptions of linearity and sparsity.

The main contribution of this work is to introduce more rigorous and quantitative ways of measuring the success of sparse coding methods applied to language models. This allows us to properly assess the extent to which activations can accurately be modelled as sparse linear combinations of feature vectors, and see how the level of sparsity depends on properties of the model. In summary, we:
\vspace{-1.5mm}
\begin{itemize}
    \item Propose novel metrics for measuring the success of sparse coding on neural network activations, and demonstrate their robustness on synthetic activations;
    \vspace{-1mm}
    \item Use our metrics to provide quantitative evidence that neural network activations can be accurately modelled as sparse linear combinations of feature vectors, supporting claims in earlier works \citep{yun2021transformer, cunningham2023sparse};
    \vspace{-1mm}
    \item Provide a more thorough analysis of the success of sparse coding as compared to previous works, including studying the relative sparsity of various model types and across layers.
\end{itemize}
\vspace{-1mm}
We provide a more thorough review of related literature in Appendix \ref{app:litreview}.

\section{Background}

Formally, the \textit{linear representation hypothesis} can be interpreted in the following way. At a fixed layer of a given neural network, we assume that the activations represent $m$ different ground-truth features $\F_1, \dots, \F_m$, each encoded by a feature vector $\f_1, \dots, \f_m \in \R^d$. An input containing features $\F_{i_1}, \dots, \F_{i_k}$ should be represented by an activation vector which is a linear combination of $\f_{i_1}, \dots, \f_{i_k}$. Conversely, any activation $\x \in \R^d$ should be approximately decomposable as a linear sum $\x \approx \alpha_{i_1} \f_{i_1} + \dots + \alpha_{i_k} \f_{i_k}$ where $\F_{i_1}, \dots, \F_{i_k}$ are the features present in the input producing activation $\x$ and $\alpha_{i_1}, \dots, \alpha_{i_k}$ are non-negative feature coefficients. We call $m$ the \textit{dictionary size} and $d$ the \textit{embedding size}. For notational convenience we combine the feature vectors into a matrix $\Phi \in \R^{d \times m}$ whose columns are $\f_1, \dots, \f_m$.

The \textit{sparsity hypothesis} says that a typical activation $\x$ only requires a small number of features to approximately represent it as a linear combination of those features, that is, we can find $\alpha_{i_1}, \dots, \alpha_{i_k}$ such that $\x \approx \alpha_{i_1} \f_{i_1} + \dots + \alpha_{i_k} \f_{i_k}$ and $k \ll d$ \citep{arora2018linear, elhage2022toy}. In practice, we expect that most features of an input should be active on only relatively few inputs, making sparsity a natural assumption \citep{arora2018linear, elhage2022toy}.

In our language model setting, we observe a set of intermediate activations $\x^{(1)}, \dots, \x^{(n)}$ from a given layer. Our task is to reconstruct the feature vectors $\f_1, \dots, \f_m$ corresponding to the ground-truth features. If the set of activations is fixed, we may combine them into a matrix $X \in \R^{d \times n}$ with columns $\x^{(1)}, \dots, \x^{(n)}$ and formulate our problem as finding $\Phi \in \R^{d \times m}$ and $\alpha \in \R^{m \times n}$ such that $X = \Phi \alpha + \eps$ where $\alpha \succeq 0$ is sparse and $\eps$ is small, typically in $L^2$ norm. (Alternatively, we may consider our activations to be drawn from some activation distribution; see Appendix \ref{app:distribution} for details.) 

For a fixed matrix $X$ of activations, the sparse coding problem is frequently solved by minimizing
\begin{equation}
\label{eq:sparsecodingobjective}
    \cal{L}(\Phi; \alpha) = \frac{1}{n} \lr{\| X - \Phi \alpha \|_2^2 + \lambda \|\alpha\|_1},
\end{equation}
which we call the the \textit{sparse coding} objective, over $\alpha \in \R^{m \times n}$ and $\Phi \in \R^{d \times m}$, where the $L^1$ norm is viewed as a continuous relaxation of the $L^0$ sparsity norm and $\lambda$ is a hyperparameter controlling the degree of sparsity regularization. We constrain the columns of $\Phi$ to have $L^2$ norm 1 and minimise $\cal{L}(\Phi)$ using an iterative optimization procedure similar to that of \citep{beck2009fast, yun2021transformer}, as described in Appendix \ref{app:algorithm}.

For our metrics defined later, it will be convenient to define $\alpha(\Phi) = \argmin_{\alpha \in \R^{m \times n}} \cal{L}(\Phi; \alpha)$, i.e. the optimal choice of $\alpha$ for a given $\Phi$, and $\cal{L}(\Phi) = \cal{L}(\Phi, \alpha(\Phi))$, i.e. the value of the sparse coding objective for a given $\Phi$, assuming an optimal choice of $\alpha$. In practice, we approximate $\alpha(\Phi)$ using the same optimization procedure described in Appendix \ref{app:algorithm} for minimizing \eqref{eq:sparsecodingobjective}.

\section{Metrics for Success of Sparse Dictionary Learning}
\label{sec:metrics}

For a sparsity metric to be meaningful, it should be invariant under scaling all activations by the same factor, continuous with respect to the activations (since we expect activations not to be an exact linear combination of feature vectors, so our metric should be robust to small perturbations) and ideally intuitive. For data which is genuinely generated using a sparse linear mechanism with an average of $k$ fully active features, we'd like our metric to be approximately $k$. We consider four metrics, the first two of which have been previously considered \citep{sharkey2022taking}, and two which are novel as far as we are aware.

\textbf{Non-zero entries:} The most natural metric of success is the average number of non-zero entries in the coefficient vector $\alpha$, which we denote $\cal{N}_0(\Phi) = \tfrac{1}{n} \|\alpha(\Phi)\|_0$. Though $\cal{N}_0(\Phi)$ is intuitive and invariant under scaling, we find that it is not robust in practice.

\textbf{Final loss value:} Second, we consider using $\cal{L}(\Phi)$ for the final value of $\Phi$ \citep{sharkey2022taking}. Though this metric is continuous, it is not invariant under scaling and so does not give useful comparisons across different models or layers within a model.

\textbf{Average coefficient norm:} Third, we consider $\cal{S}_p(\Phi) = \|\alpha(\Phi)\|_p^p / \|\alpha(\Phi)\|_\infty^p$ for each $p > 0$. This corresponds to the average $L^p$ norm of the coefficient vector, normalized by the average maximum coefficient. This is scale invariant and, in the case where all feature coefficients are either zero equal to the same positive value, $\cal{S}_p(\Phi)$ corresponds to the average number of features present. In practice, we typically use $p=1$ for simplicity and robustness. 

\textbf{Normalized loss:} We define the normalized loss as $\cal{L}_{\textup{norm}}(\Phi) = \cal{L}(\Phi) / (\lambda \|\alpha(\Phi)\|_\infty)$. This is similar to the final loss value, but normalized by the average magnitude of the feature coefficients, restoring invariance to scaling. Intuitively, if we perfectly reconstruct the activations so that $\x = \Phi \alpha$ always, then $\cal{L}(\Phi) = \lambda \|\alpha(\Phi)\|_1$ and so $\cal{L}_{\textup{norm}}(\Phi) = \cal{S}_1(\Phi)$. In practice, reconstruction is somewhat imperfect, in which case $\cal{L}_{\textup{norm}}$ also includes a penalty for the unreconstructed part. This makes it more stable over a range of hyperparameters.

\subsection{Experimental Verification of Metrics}
\label{sec:metric_verification}

We now test the effectiveness of the four proposed metrics by evaluating them on various synthetic datasets where we know the true level of sparsity. First, we test how well our metrics can predict the true level of sparsity for sparse linear data. To do this, we generate synthetic activation distributions that satisfy the sparsity and linearity hypotheses with varying average numbers $a$ of active features. Details of the generating process are given in Appendix \ref{app:syntheticdata}. We then decompose our synthetic activations using sparse coding and observe how well our metrics predict the true sparsity.

\begin{figure}[b]
    \centering
    \begin{subfigure}[b]{0.6\textwidth}
        \centering
        \raisebox{1.3cm}{\includegraphics[width=\textwidth]{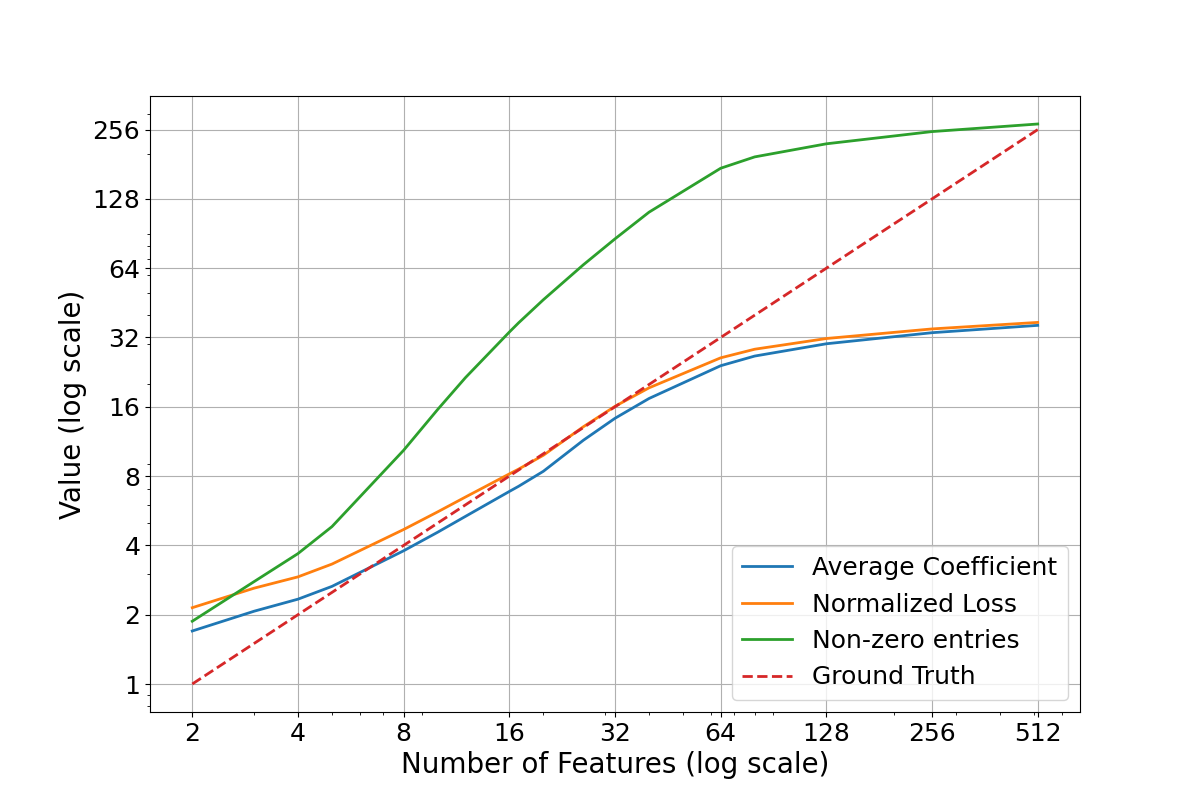}}
    \end{subfigure}
    \begin{subfigure}[b]{0.3\textwidth}
        \centering
        \includegraphics[width=\textwidth, trim={0cm 0cm 18cm 15cm},clip]{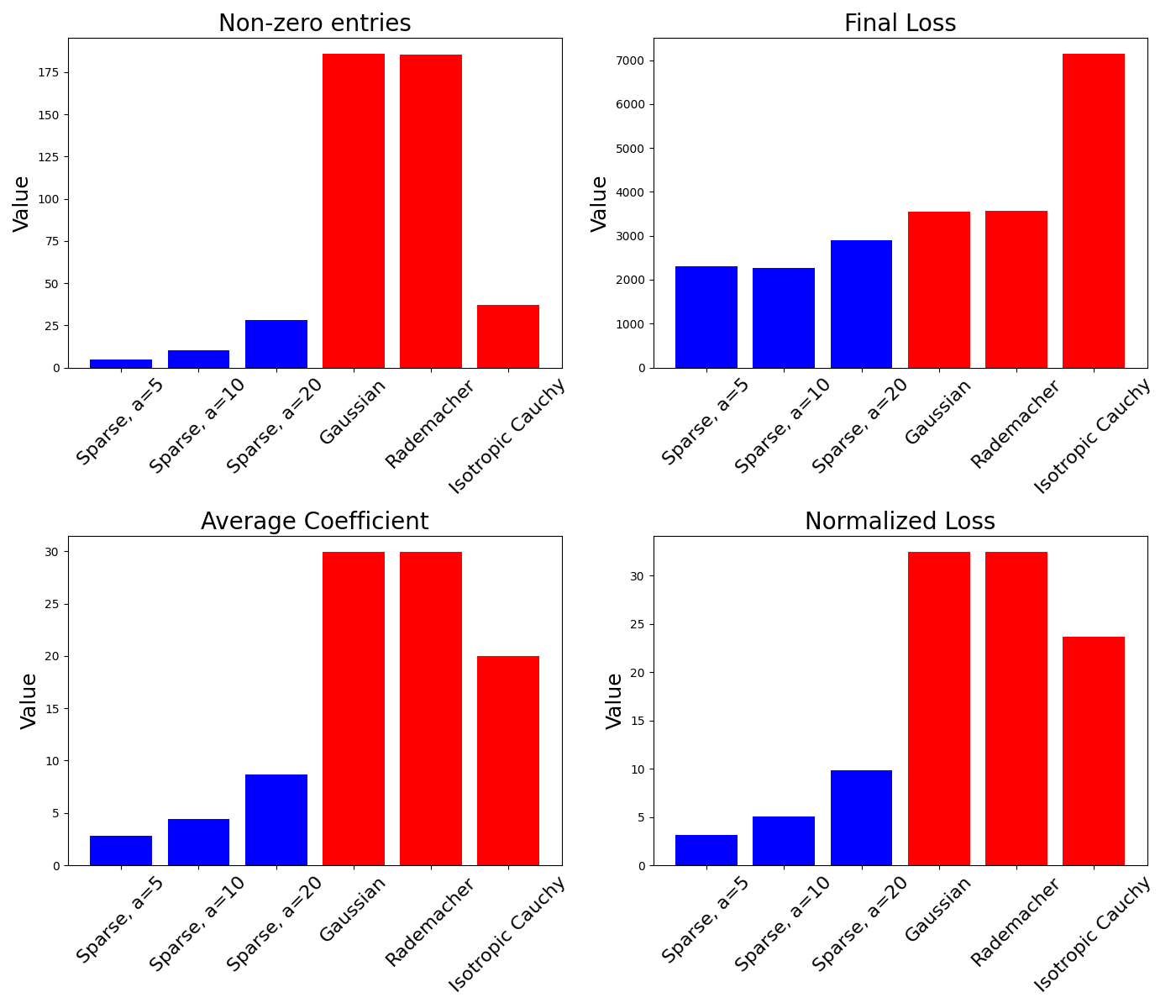}
        \includegraphics[width=\textwidth, trim={18cm 0cm 0cm 15cm},clip]{images/datasets_4feats_8Guess_0.1Noise_256Dims.png}
    \end{subfigure}
    \caption{(a) Metric values compared to true sparsity level for synthetic data. (b) Metric values for sparse linear data (blue) compared to non-sparse linear data (red).}
    \label{fig:truevsactualsparsity}
\end{figure}

We plot the true sparsity $a$ against each metric in \cref{fig:truevsactualsparsity}(a). We see that for metric values less than 20, the normalized loss and average coefficient norm closely approximate the true sparsity, while plateauing for greater values. This suggests that these metrics can reliably approximate the correct sparsity level for low sparsities. Meanwhile, number of non-zero features overestimates the true level of sparsity. The final loss is not scale-invariant, so it cannot be compared to the true sparsity.

Second, we test how well our metrics can distinguish between sparse linear data and other datasets. We construct three sparse linear datasets and three non-sparse linear datasets (details in Appendix \ref{app:syntheticdata}), decompose each of the datasets using sparse coding and apply our metrics. The results for average coefficient and normalized loss are shown in \cref{fig:truevsactualsparsity}(b). We see that the average coefficient norm and normalized loss perform very similarly, and both clearly distinguish between the sparse linear data and the other datasets. The results for non-zero entries and final loss are shown in Appendix \ref{app:syntheticdata}; both perform less well and so we focus on average coefficient norm and normalized loss from now on.

We also consider ablations of both experiments where we vary embedding size, dictionary size, noise level and number of ground truth features. We find our results are robust to changes in noise level, dictionary size and number of ground-truth features, and relatively robust to changes in embedding size, provided the level of sparsity is not too close to the embedding size. Performance may deteriorate if the number of ground-truth features is too much larger than the embedding dimension, since then our linear sparse data starts to look approximately Gaussian. For details, see Appendix \ref{app:ablationsmetricverification}.

\section{Demonstrating Sparsity in Language Model Activations}
\label{sec:llms}

Now that we have demonstrated that average coefficient norm and normalized loss can reliably distinguish between sparse linear activations and some other classes of distributions, we use them to study sparsity in language model activations. In this section, we focus on the normalized loss, as it is more robust in practice; we present results using average coefficient norm in the appendices.

\subsection{Embedding Layers}
\label{sec:embeddinglayers}

First, we use our metrics to assess sparsity in the embedding layers of transformer language models. We pick three classes of models to test on: BERT (Tiny, Mini, Small and Medium) \citep{turc2019well, bhargava2021generalization}, TinyStories (1M, 3M and 33M) \citep{eldan2023tinystories}, and GPT-Neo/GPT-2 \citep{black2021gptneo, radford2019language}. We use the token embeddings as our set of activations $X$, apply sparse coding, and measure the sparsity of the resulting decomposition using normalized loss. See Appendix \ref{app:llmsfurtherdetails} for experimental details and results for average coefficient.

\begin{figure}[b]
    \centering
    \includegraphics[width=0.55\textwidth]{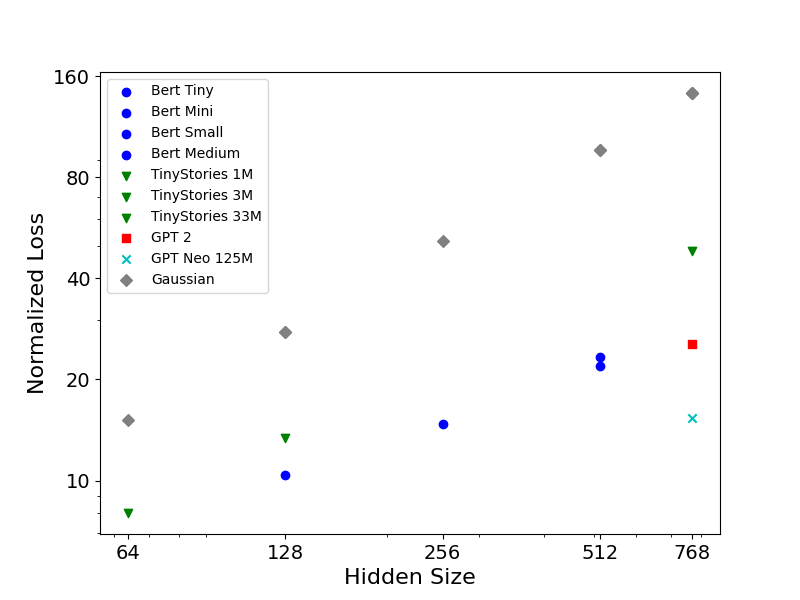}
    \caption{Embedding size versus normalized loss for token embeddings of three different classes of language models.}
    \label{fig:embdding-sparse}
\end{figure}

The normalized loss values we obtain are displayed in \cref{fig:embdding-sparse}, where we have also plotted the sparsity value achieved for a standard Gaussian distribution for reference. We observe that for all models the normalized loss of the token embeddings is much lower than that for a Gaussian, as we would expect if the linearity and sparsity hypotheses held for the embeddings.

Note also that if we compare the normalized loss values in \cref{fig:embdding-sparse} to the results in Section \ref{sec:metric_verification} and Appendix \ref{app:ablationsmetricverification}, we see they are within the range where normalized loss correctly predicts ground-truth sparsity on synthetic sparse linear data. This suggests that the normalized loss values in \cref{fig:embdding-sparse} are a good approximation to the true sparsity level, assuming the linearity and sparsity hypotheses.

Second, we observe that the number of active features increases as the embedding size increases, but more slowly. This matches the intuition that the model can disentangle more semantic meanings in a higher dimensional space. We also observe that the BERT architecture leads to slightly more sparse representations than the TinyStories architecture.

As observed in previous work, the feature decompositions produced by our sparse coding method do frequently correspond to natural human interpretations. In Appendix \ref{app:maxactivatingembedding}, we provide examples where our feature decomposition splits tokens into multiple semantically meaningful components, and indicate that the feature directions that we find may be more interpretable than arbitrary directions in embedding space, corroborating the findings of \citet{yun2021transformer} and \citet{cunningham2023sparse}.

\subsection{Later Layers}
\label{sec:laterlayers}

Next, we explore how the level of sparsity changes across layers of a language model. We apply our sparse coding method to all layers of four BERT models (Tiny, Mini, Small and Medium), with activations generated by running the models on Wikipedia abstracts, and plot the resulting normalized loss metric. See Appendix \ref{app:llmlaterlayers} for experimental details and results using average coefficient.

\begin{figure}[t!]
    \centering
    \begin{subfigure}[b]{0.48\textwidth}
        \centering
        \includegraphics[width=\textwidth]{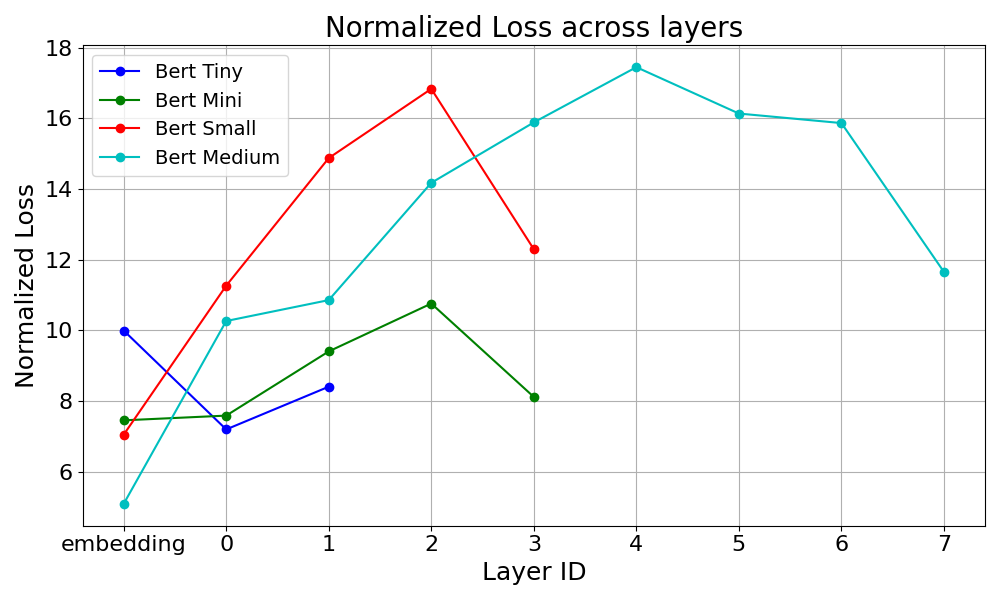}
    \end{subfigure}
    \begin{subfigure}[b]{0.48\textwidth}
        \centering
        \includegraphics[width=\textwidth]{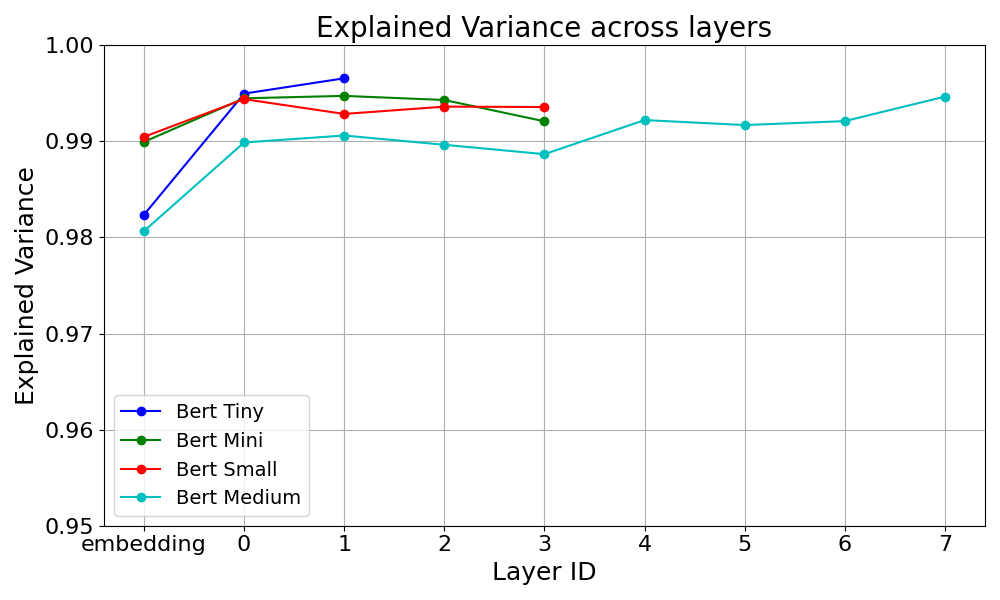}
    \end{subfigure}
    \caption{(a) Sparsity of activations by layer for four BERT models. (b) Percentage of activation variance explained by our sparse decomposition for each layer and model.}
    \label{fig:sparsitybylayer}
\end{figure}

The results are plotted in \cref{fig:sparsitybylayer}. We find that all layers are sparse according to the normalized loss. In addition, there is a general trend that the embedding layer is the most sparse, with sparsity first decreasing as layer depth increases, before becoming more sparse again in the latest layers. This is consistent with the intuition that the model may initially be detecting a larger number of higher-level features in deeper layers, leading to a reduction in sparsity as layer index increases, before having to extract just the key features for the next token prediction, leading to a subsequent decrease in features stored in the final layers and corresponding increase in the sparsity level.

\section{Social Impacts}

In this work, we have developed novel metrics for measuring the success of sparse coding, we have demonstrated that our metrics can accurately predict the level of sparsity for synthetic models of sparse linear activations, and we have applied these metrics to measure sparsity in language models. We show that activations across various model architectures and all layers were more sparse than several control datasets, providing evidence for the linearity and sparsity hypotheses, and find that activation sparsity is greatest in the first and final layers of a model.

Increasing the interpretability and transparency of language models is a central tool for ensuring that their deployment is safe and broadly beneficial. Determining the features that these models are relying on to make their predictions is a key step in rendering them interpretable. Sparse coding is one particularly promising avenue for feature extraction, since it has demonstrated initial success and runs automatically and so has the potential to scale well to large models.

To improve sparse coding for feature extraction, it is important to be able to reliably assess the performance of current methods and the accuracy of the assumptions on which they rely. We hope that by providing better quantitative ways to assess feature sparsity and superposition within language models, we can encourage more detailed and precise studies of these methods, including more rigorous tests of the linearity and sparsity hypotheses, further investigations into the factors that affect the degree of superposition within models, and improved methods to disentangle superposed features.


\bibliography{iclr2024_conference}
\bibliographystyle{abbrvnat}

\newpage

\appendix

\section{Related Works}
\label{app:litreview}

\textbf{Linear representations}\;\; There is considerable evidence that neural networks learn representations with linear structure: for example, linear operations on Word2Vec embeddings capture semantic meaning \citep{mikolov2013linguistic, phamvan2020interpreting}, and linear interpolation in the latent space of GANs and VAEs can combine the features of multiple datapoints \citep{bojanowski2018optimizing, berthelot2019understanding}. In language models, several works have demonstrated success using linear techniques for locating locating information within models \citep{meng2022locating, burns2023discovering} and editing model behaviors \citep{ilharco2022editing, ravfogel2022linear, turner2023activation}, as well as observing linear representations in reverse-engineered circuits \citep{nanda2023progress}. Such observations motivated the introduction of the linear representation hypothesis in \cite{elhage2022toy}.

\textbf{Feature extraction}\;\; Methods proposed for discovering the features used by neural networks include feature visualization \citep{erhan2009visualizing, olah2017feature, olah2020zoom}, saliency maps \citep{simonyan2014deep}, feature importance scores \citep{lundberg2017unified}, and layer-wise relevance propagation \citep{bach2015pixel}. For transformer language models, common techniques include visualizing attention patterns \citep{wang2023interpretability, bills2023language}, gradient-based contrastive explanations \citep{yin2022interpreting}, and more recently methods based on sparse coding (see below).

\textbf{Superposition and polysemanticity}\;\; The idea that text embeddings can be modelled as a linear superposition of sparse feature vectors was developed by \citet{faruqui2015sparse} and \citet{arora2018linear}. \citet{olah2020zoom} introduced the notion of polysemanticity in the context of vision models, and hypothesised that polysemantic neurons arise to allow networks to represent more features with a fixed number of neurons. The first theoretical models for polysemanticity and superposition were presented by \citet{elhage2022toy} and \citet{scherlis2022polysemanticity}.

\textbf{Sparse coding}\;\; The sparse coding problem was introduced by \cite{olshausen1996emergence, olshausen1997sparse}. A variety of methods for learning the sparse dictionary have been proposed, including the method of optimal directions (MOD) \citep{engan1999method}, $k$-SVD \citep{aharon2006ksvd}, and methods using Lagrange duality \citep{lee2006efficient}. Several works have applied sparse coding to language models using techniques based on autoencoders \citep{sharkey2022taking, cunningham2023sparse}, or FISTA \citep{beck2009fast,yun2021transformer}. The method we use in this paper is a variant of the iterative optimization method used by \cite{yun2021transformer}.

\textbf{Automated interpretability}\;\; Recently proposed methods for automating the interpretation of neural networks include using multimodal models to automatically propose interpretations for neurons based on activation patters \citep{oikarinen2023clip, bills2023language}, using causal scrubbing to automatically detect circuits within language models \citep{chan2022causal, conmy2023towards}, and using singular value decompositions of weight matrices \citep{millidge2022singular}. The previous works most closely related to our current contributions are \citet{yun2021transformer} and \citet{cunningham2023sparse}, who both use sparse coding to automatically identify a set of feature directions in activation space. They both measure the success of their methods by inspecting sets of examples where a particular feature is active and aiming to identify the common feature between the examples, using either a human labeller or a language model.

\section{Setup for an Activation Distribution}
\label{app:distribution}

In practice, we can typically generate new intermediate activations at will by running our model on new text. Accordingly, it can make sense to imagine that we have an activation distribution from which we can sample, rather than a fixed finite set of given activations $\x^{(1)}, \dots, \x^{(n)}$. With this setup, our problem amounts to finding $\Phi \in \R^{d \times m}$ such that for an activation $\x$ drawn from this distribution we can find $\alpha \in \R^m$ and $\eps$ such that $\x = \Phi \alpha + \eps$, where $\alpha$ is typically sparse and $\eps$ is small in expectation.

For convenience, in this paper we will stick to notating the case of a fixed matrix $X$ of activations, but all quantities discussed can be trivially extended to the case where we draw samples $\x$ from the distribution of activations -- wherever we sum over rows of $X$ or $\alpha$, this should instead be replaced by an expectation over the activation distribution.

\section{Details of Sparse Coding Algorithm}
\label{app:algorithm}

The algorithm used in this paper for minimizing the sparse coding objective $\cal{L}(\Phi; \alpha)$ from \eqref{eq:sparsecodingobjective} over $\alpha \in \R^{m \times n}$ and $\Phi \in \R^{d \times m}$ is as follows. It is inspired by the iterative optimization procedure used by \citet{yun2021transformer}.

We alternate between the following two steps. First, we minimize the objective $\cal{L}(\Phi; \alpha)$ over all $\Phi \in \R^{d \times m}$ using several steps of stochastic gradient descent. Second, we update $\alpha$ using a greedy procedure. For each column $\x^{(j)}$ of $X$, we find the current feature vector $\f_i$ that has the largest dot product with $\x$, say $\f_{i_1}$, and set $\alpha_{i_1, j} = \x \cdot \f_{i_1} - \tfrac{1}{2} \lambda$ (this choice corresponds to projection under the $L^1$ norm). If $\alpha_{i_1, j} > 0$, then we subtract $\alpha_{i_1, j} \f_{i_1}$ from $\x^{(j)}$, find the feature vector with the next largest dot product and repeat. We do this until $\alpha_{i_k, j} \leq 0$ at some step, at which point we set all remaining $\alpha_{s,j} = 0$. We repeat this for each column of $X$ to find the new matrix $\alpha \in \R^{m \times n}$.

We alternate these two steps, updating $\Phi$ and $\alpha$ sequentially until the process converges. In the case where the activations are drawn from a distribution rather than given by a fixed matrix $X$, we resample the activations at each iteration.

We find empirically that the best choice of $\lambda$ is approximately 10\% of the maximum activation, i.e. $\lambda \approx 0.1 \|\alpha\|_\infty$. Where it is computationally feasible, we use an adaptive scheme to set $\lambda$, first guessing a reasonable choice of $\lambda$, then running our sparse coding procedure once to get a feature coefficient matrix $\alpha$, then updating $\lambda = 0.1 \|\alpha\|_\infty$ and iterating until convergence. In practice, we find that we typically only need a couple of steps to converge.

\section{Further Details on Metric Verification Experiments}
\label{app:syntheticdata}

To generate synthetic sparse linear data with $a$ features active on average, we do the following. First, we generate $m = 4d$ feature vectors $\f_1, \dots, \f_m \in \R^d$ sampled uniformly from the unit sphere. Then, to generate each activation we sample a vector $\alpha$ of feature coefficients by taking each coefficient to be uniform between $0$ and $1$ with probability $a/4d$ and zero otherwise. We define the proto-activation to be $\hat \x = \sum_i \alpha_i \f_i$, so that on average each proto-activation is the sum of $a$ activated vectors. We center the set of proto-activations and finally add Gaussian noise of variance $a \sigma^2 / d$ to each proto-activation to get our synthetic activations.

Note that since each active feature is weighted by a factor distributed uniformly between $0$ and $1$, the expected weighted number of active features will be $a/2$. Hence, we consider the ``correct'' value of our metric on this dataset to be $a/2$ (rather than $a$).

In our initial experiments in Section \ref{sec:metric_verification} we take $d = 256$, $\sigma = 0.1$ and use $16384$ datapoints and a dictionary size of $8d$. We consider the effects of varying the embedding size, noise level, number of ground-truth features and dictionary size in Appendix \ref{app:ablationsmetricverification}. 

For our three non-sparse linear datasets, we use (i) a Gaussian distribution with identity covariance (not sparse), (ii) an heavy-tailed isotropic distribution constructed by sampling from an isotropic Gaussian and then scaling $\|\x\|_2$ to be Cauchy distributed (heavy-tailed but not sparse), and (iii) a $d$-dimensional Rademacher distribution (sparse but not well-represented by a sparse linear combination of features). Each distribution is scaled to have identity covariance.

\begin{figure}[b!]
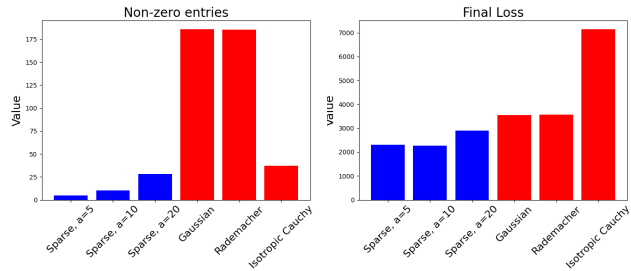

    \centering
    \begin{subfigure}[b]{0.3\textwidth}
        \centering
        \includegraphics[width=\textwidth, trim={0cm 15.2cm 18cm 0cm},clip]{images/datasets_4feats_8Guess_0.1Noise_256Dims.png}
    \end{subfigure}
    \begin{subfigure}[b]{0.3\textwidth}
        \centering
        \includegraphics[width=\textwidth, trim={18cm 15.2cm 0cm 0cm},clip]{images/datasets_4feats_8Guess_0.1Noise_256Dims.png}
    \end{subfigure}
    \caption{Metric values for sparse linear data (blue) compared to non-sparse linear data (red), with non-zero entries and final loss metrics.}
    \label{fig:truevsactualsparsityothermetrics}
\end{figure}

\cref{fig:truevsactualsparsityothermetrics} shows the results of attempting to use non-zero entries and final loss to distinguish between sparse linear data and other datasets. We see that non-zero entries fails to reliably distinguish the heavy-tailed but non-sparse data, while the final loss fails to reliably distinguish the Gaussian data from the sparse linear data. In order to make the final loss values comparable across datasets, we scale all datasets to have mean $0$ and the average $L_2$ norm equal $1$.

\section{Ablations for Metric Verification Experiments}
\label{app:ablationsmetricverification}

In this section we assess the robustness of the results from Section \ref{sec:metric_verification} to changes in the parameters of the problem definition. We consider changing the level of noise in the construction of our synthetic dataset $\sigma$, the size of the embedding dimension $d$, the dictionary size, and the number of ground-truth features.

\begin{figure}[t!]
    \centering
    \begin{subfigure}[b]{0.6\textwidth}
        \centering
        \raisebox{1.3cm}{\includegraphics[width=\textwidth]{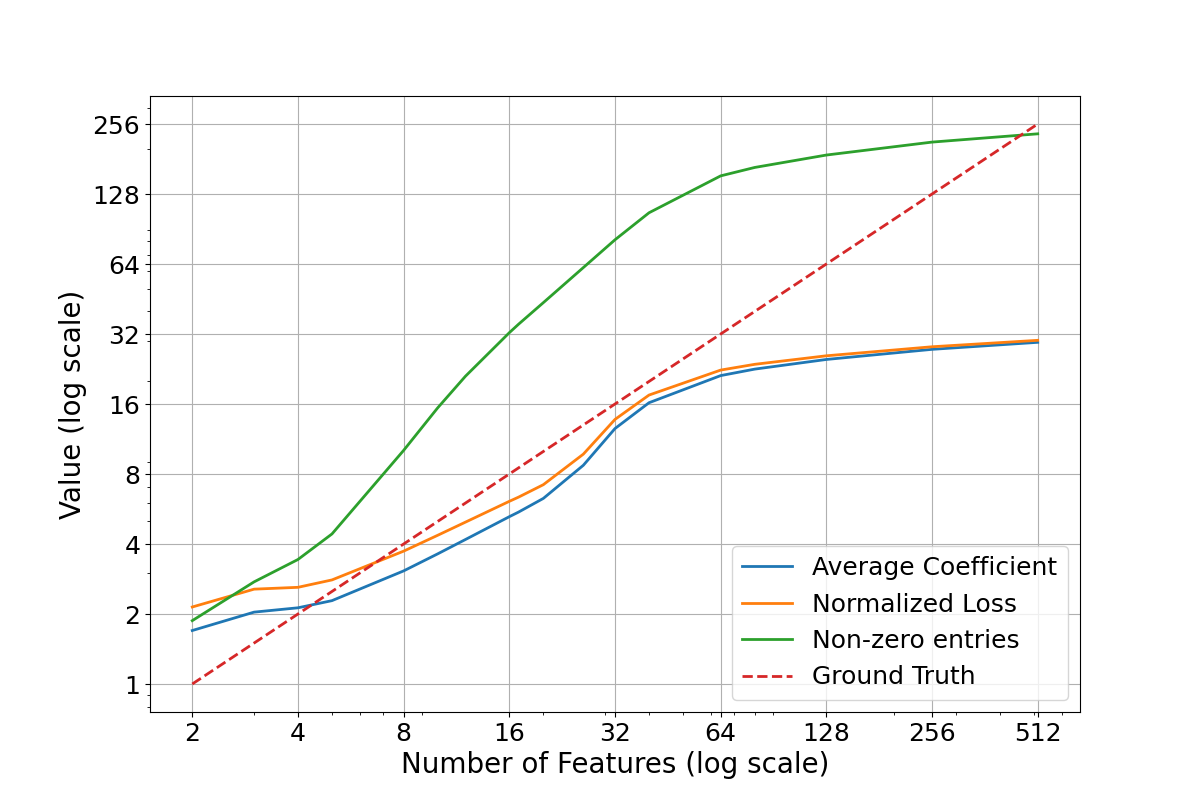}}
    \end{subfigure}
    \begin{subfigure}[b]{0.3\textwidth}
        \centering
        \includegraphics[width=\textwidth, trim={0cm 0cm 18cm 15cm},clip]{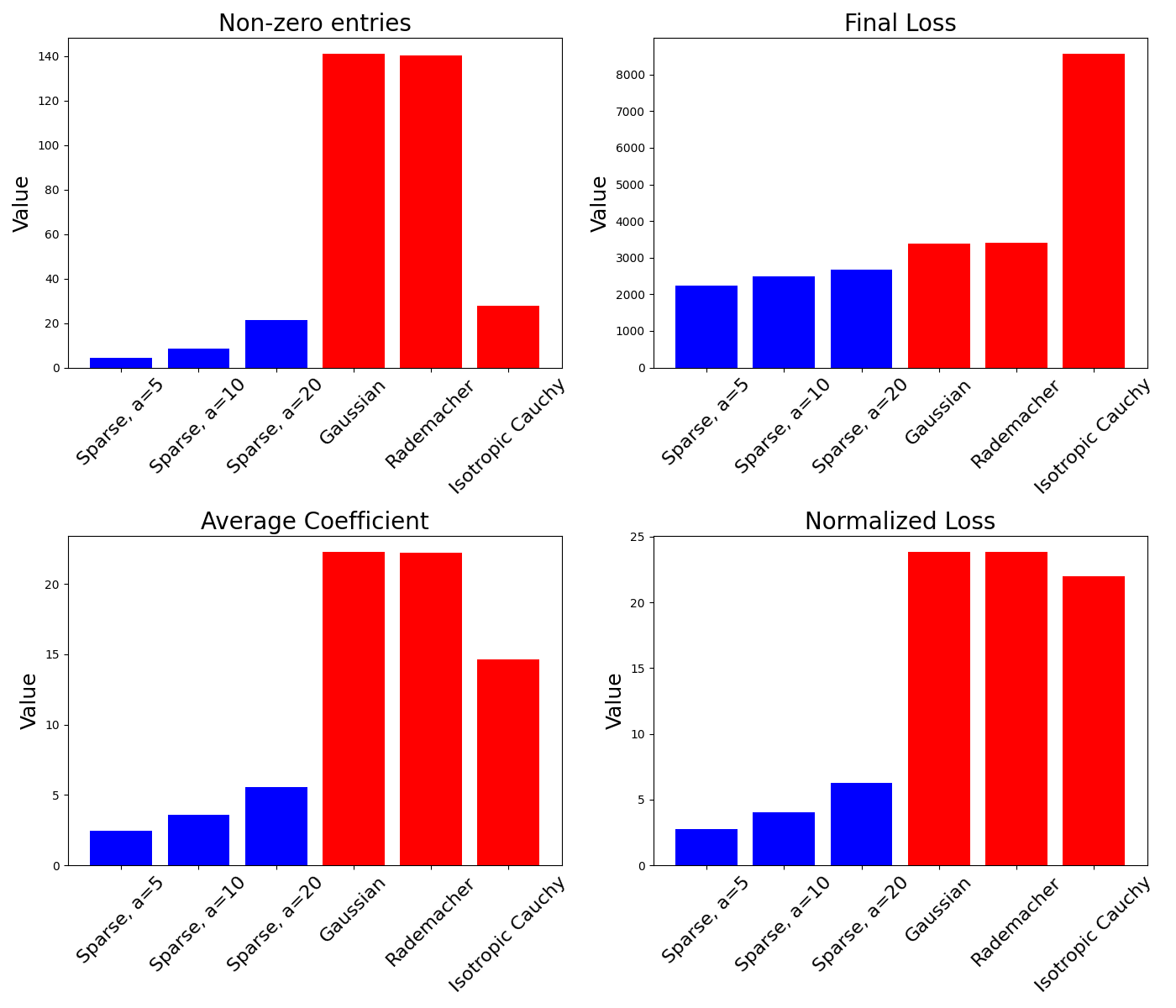}
        \includegraphics[width=\textwidth, trim={18cm 0cm 0cm 15cm},clip]{images/datasets_4feats_16Guess_0.1Noise_256Dims.png}
    \end{subfigure}
    \caption{Results of experiment from Section \ref{sec:metric_verification} with dictionary size of $16d$. (a) Metric values compared to true sparsity level for synthetic data. (b) Metric values for sparse linear data (blue) compared to non-sparse linear data (red).}
    \label{fig:dictionarysizeablation}
\end{figure}

First, we assess the effect of changing the dictionary size. We keep the same embedding dimension $d = 256$, the same number of ground-truth features $4d$ and the same noise level $\sigma = 0.1$ as in the main text, but consider increasing the dictionary size to $16d$. We plot the results of this experiment in \cref{fig:dictionarysizeablation}. We see that the results are very similar to \cref{fig:truevsactualsparsity}; the average coefficient and normalized loss metrics continue to distinguish well between the sparse linear data and the other datasets, and both metrics track the ground-truth sparsity well for metric values below approximately 16. We conclude that our methods are relatively robust to dictionary sizes that are misspecified up to a factor of at least 4.

\begin{figure}[bt]
    \centering
    \begin{subfigure}[b]{0.6\textwidth}
        \centering
        \raisebox{1.3cm}{\includegraphics[width=\textwidth]{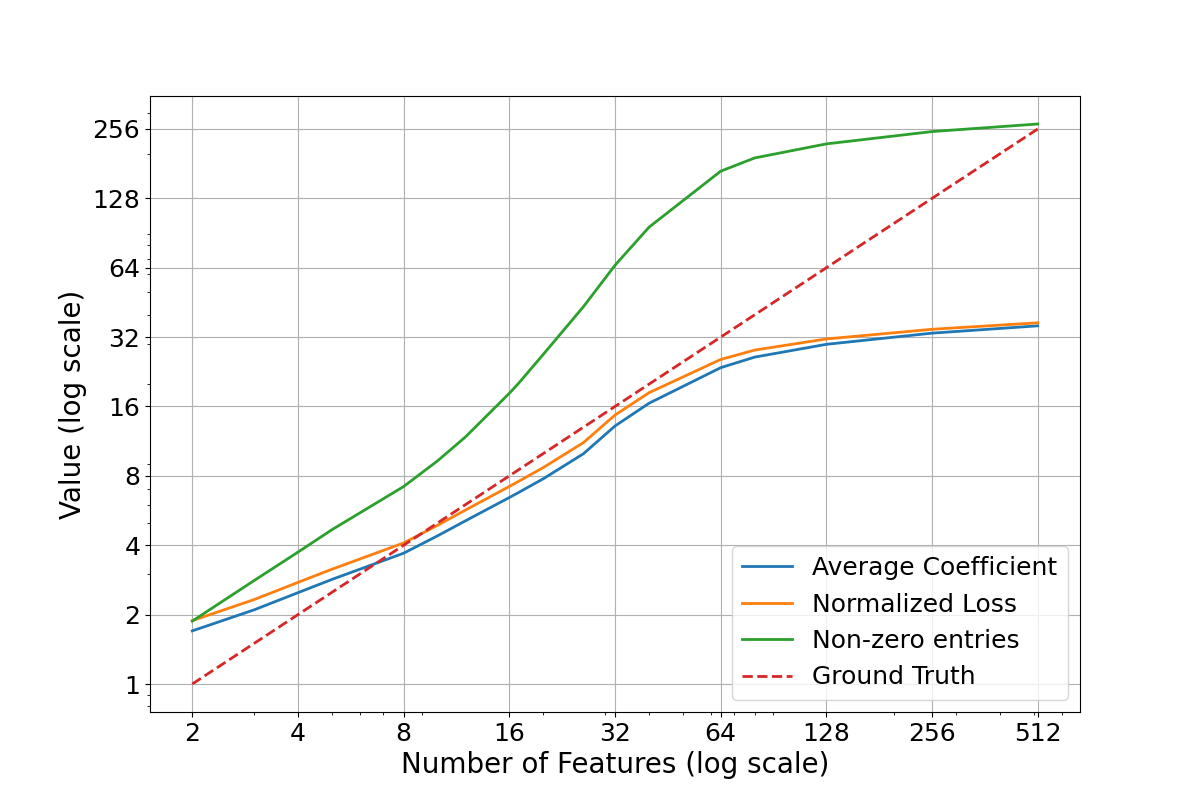}}
    \end{subfigure}
    \begin{subfigure}[b]{0.3\textwidth}
        \centering
        \includegraphics[width=\textwidth, trim={0cm 0cm 18cm 15cm},clip]{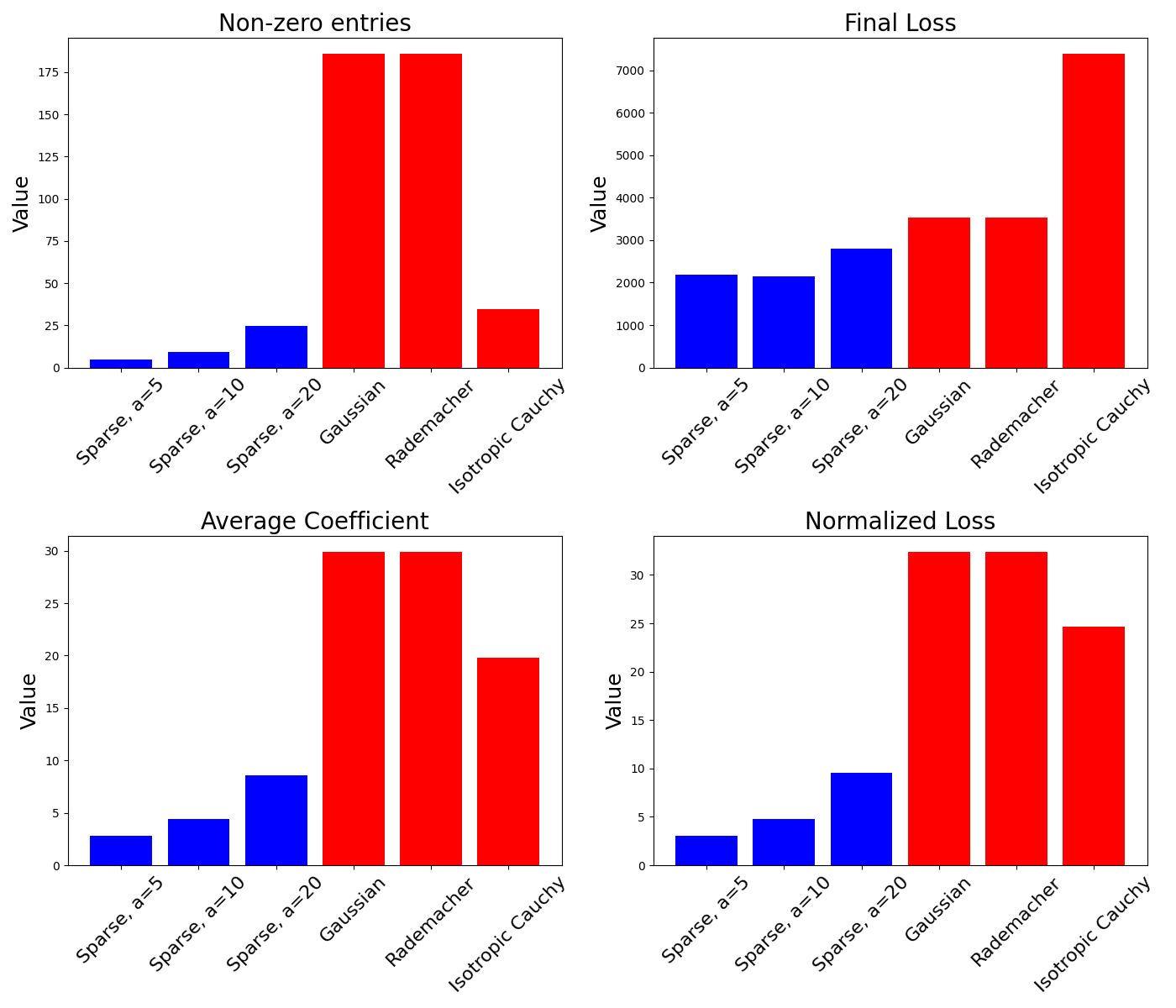}
        \includegraphics[width=\textwidth, trim={18cm 0cm 0cm 15cm},clip]{images/datasets_4feats_8Guess_0.05Noise_256Dims.png}
    \end{subfigure}
    \begin{subfigure}[b]{0.6\textwidth}
        \centering
        \raisebox{1.3cm}{\includegraphics[width=\textwidth]{images/exp2_4feats_8Guess_0.05Noise_256Dims.png}}
    \end{subfigure}
    \begin{subfigure}[b]{0.3\textwidth}
        \centering
        \includegraphics[width=\textwidth, trim={0cm 0cm 18cm 15cm},clip]{images/datasets_4feats_8Guess_0.05Noise_256Dims.png}
        \includegraphics[width=\textwidth, trim={18cm 0cm 0cm 15cm},clip]{images/datasets_4feats_8Guess_0.05Noise_256Dims.png}
    \end{subfigure}
    \caption{Results of experiment from Section \ref{sec:metric_verification} with different noise levels: $\sigma = 0.05$ (top) and $\sigma = 0.2$ (bottom). (Left) Metric values compared to true sparsity level for synthetic data. (Right) Metric values for sparse linear data (blue) compared to non-sparse linear data (red).}
    \label{fig:noiseablation}
\end{figure}

Second, we assess the effect that changing the noise level $\sigma$ has. We keep the same embedding dimension $d=256$, the same dictionary size of $8d$ and the same number of ground-truth features at $4d$ as in the main text, but consider decreasing $\sigma$ to $0.05$ or increasing it to $0.2$. The results are plotted in \cref{fig:noiseablation}. We see that both the average coefficient and the normalized loss metric continue to distinguish well between sparse linear and other data, and both metrics track the ground-truth sparsity well up to a metric value of about 20.

\begin{figure}[bt]
    \centering
    \begin{subfigure}[b]{0.6\textwidth}
        \centering
        \raisebox{1.3cm}{\includegraphics[width=\textwidth]{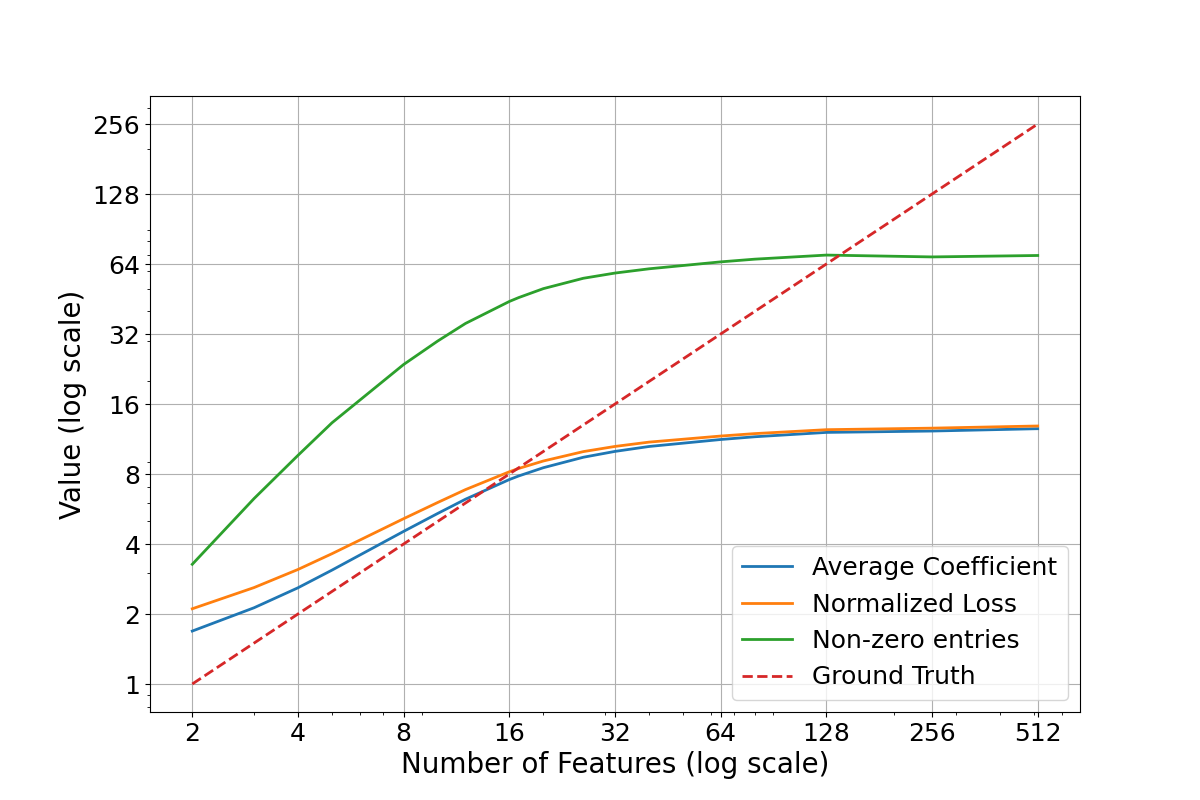}}
    \end{subfigure}
    \begin{subfigure}[b]{0.3\textwidth}
        \centering
        \includegraphics[width=\textwidth, trim={0cm 0cm 18cm 15cm},clip]{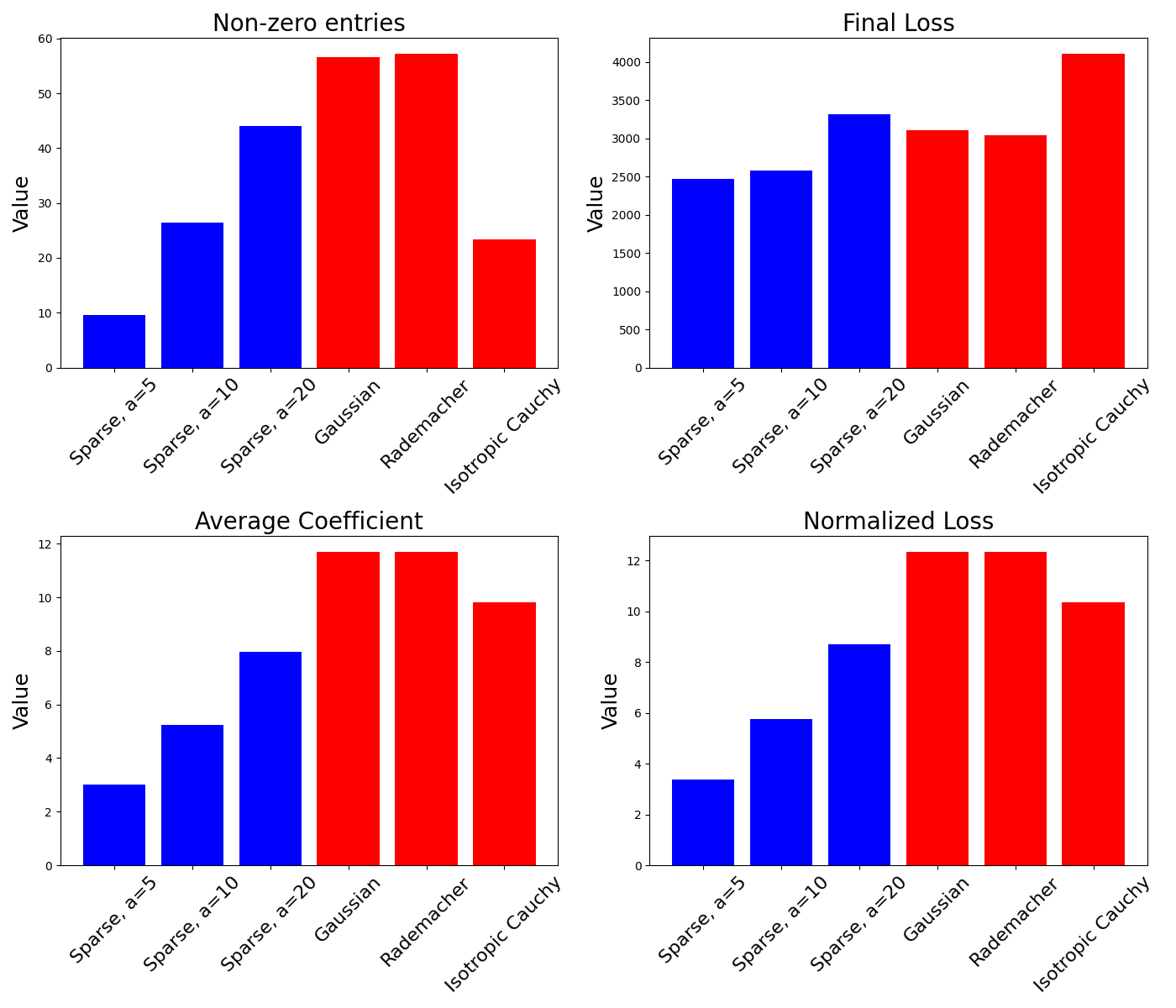}
        \includegraphics[width=\textwidth, trim={18cm 0cm 0cm 15cm},clip]{images/datasets_4feats_8Guess_0.1Noise_64Dims.png}
    \end{subfigure}
    \begin{subfigure}[b]{0.6\textwidth}
        \centering
        \raisebox{1.3cm}{\includegraphics[width=\textwidth]{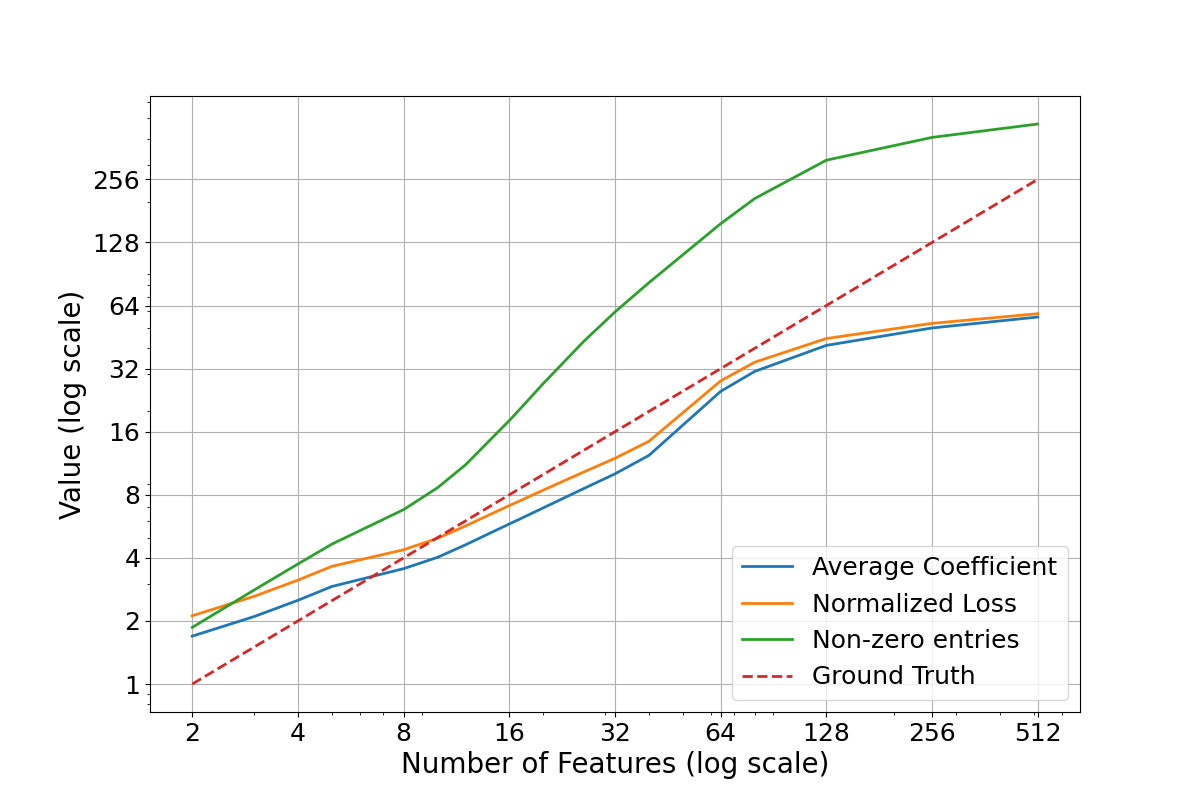}}
    \end{subfigure}
    \begin{subfigure}[b]{0.3\textwidth}
        \centering
        \includegraphics[width=\textwidth, trim={0cm 0cm 18cm 15cm},clip]{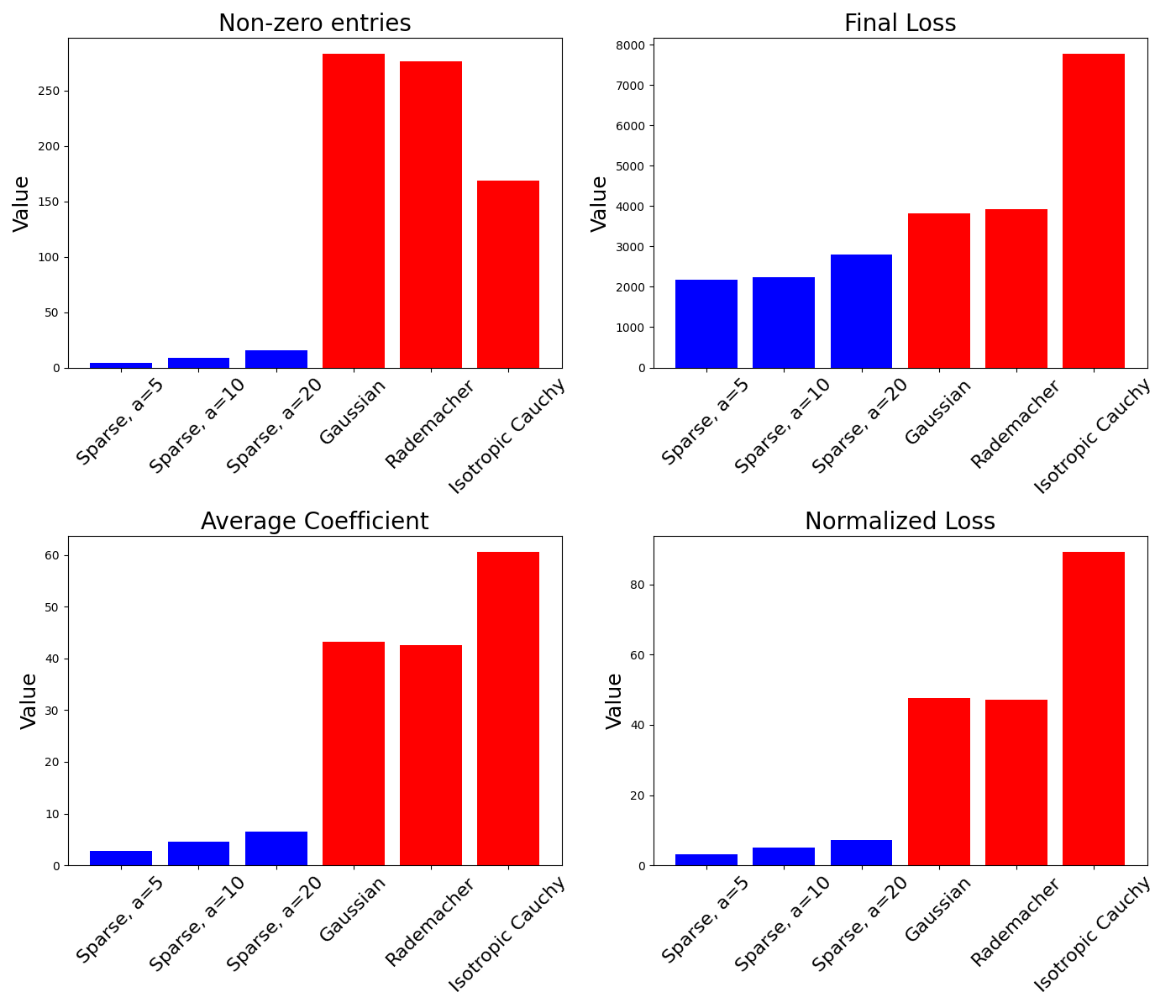}
        \includegraphics[width=\textwidth, trim={18cm 0cm 0cm 15cm},clip]{images/datasets_4feats_8Guess_0.1Noise_512Dims.png}
    \end{subfigure}
    \caption{Results of experiment from Section \ref{sec:metric_verification} with different embedding sizes: 64 (top) and 512 (bottom). (Left) Metric values compared to true sparsity level for synthetic data. (Right) Metric values for sparse linear data (blue) compared to non-sparse linear data (red).}
    \label{fig:embeddingsizeablation}
\end{figure}

Third, we assess the effect of changing the embedding size. We fix the dictionary size, number of ground-truth features and noise level as in Section \ref{sec:metric_verification} but consider taking the embedding size to be either 64 or 512. The results are shown in \cref{fig:embeddingsizeablation}. We continue to achieve good results for larger embedding sizes. For smaller embedding sizes, our metrics are less able to separate data with $a = 20$ and non-sparse data; this is likely because 20 features is a considerable fraction of the total embedding size when $d = 64$, and so data with $a = 20$ is approaching no longer being appreciably sparse. We see that our metrics still separate linear sparse data with $a = 5, 10$ and the other data sets reasonable well.

\begin{figure}[tb]
    \centering
    \begin{subfigure}[b]{0.6\textwidth}
        \centering
        \raisebox{1.3cm}{\includegraphics[width=\textwidth]{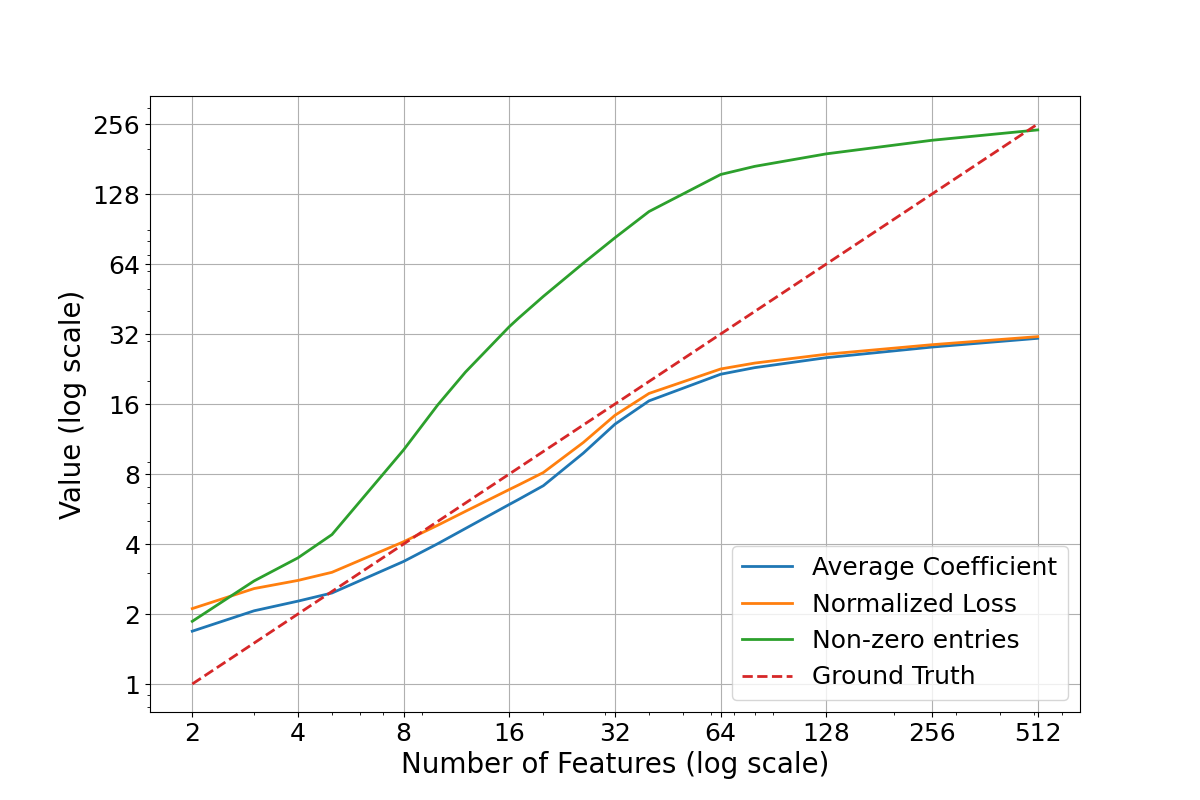}}
    \end{subfigure}
    \begin{subfigure}[b]{0.3\textwidth}
        \centering
        \includegraphics[width=\textwidth, trim={0cm 0cm 18cm 15cm},clip]{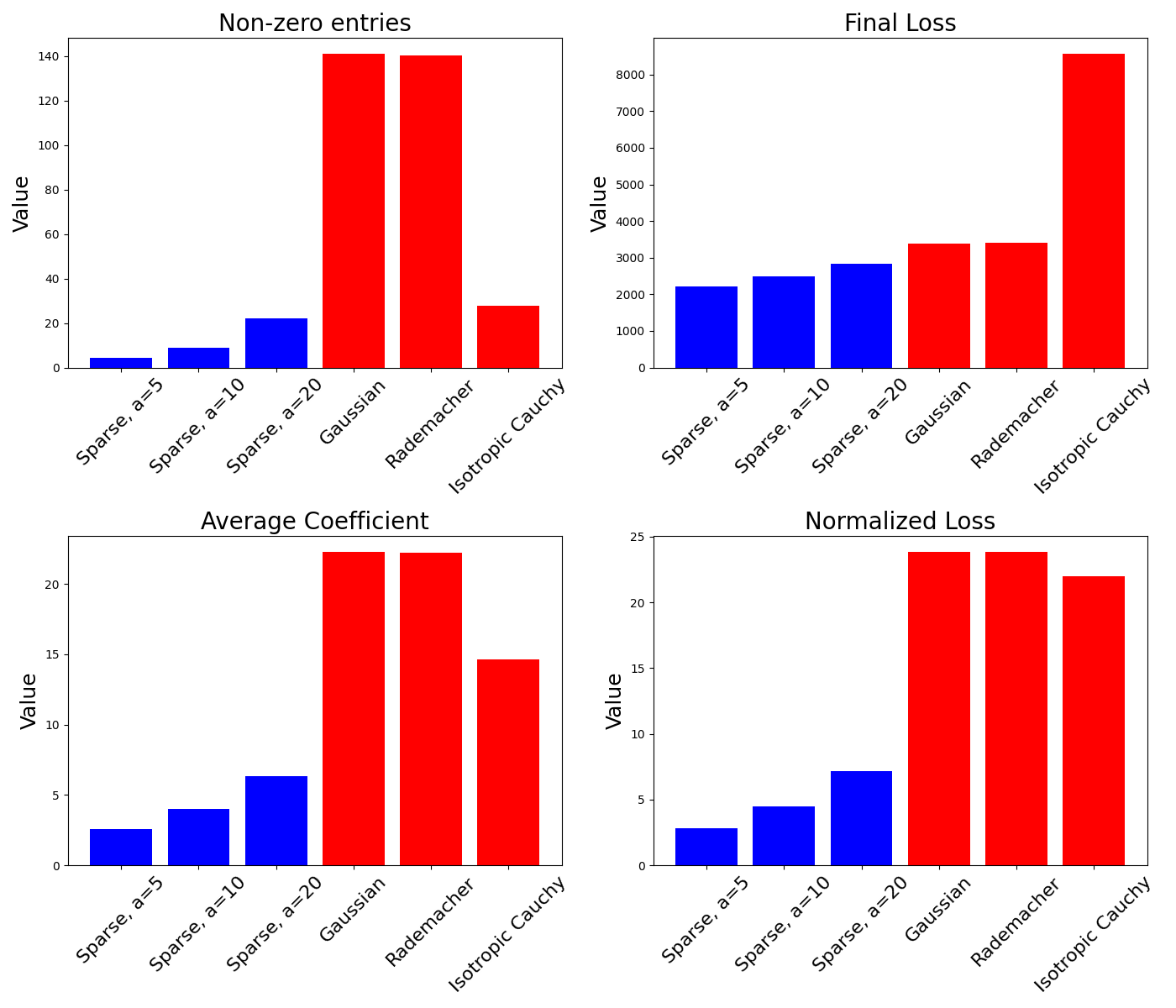}
        \includegraphics[width=\textwidth, trim={18cm 0cm 0cm 15cm},clip]{images/datasets_8feats_16Guess_0.1Noise_256Dims.png}
    \end{subfigure}
    \caption{Results of experiment from Section \ref{sec:metric_verification} with $8d$ ground-truth features. (a) Metric values compared to true sparsity level for synthetic data. (b) Metric values for sparse linear data (blue) compared to non-sparse linear data (red).}
    \label{fig:groundtruthablation}
\end{figure}

Finally, we consider changing the number of ground-truth features from $4d$ to $8d$ (while keeping all other parameters of the synthetic data the same). The results of the experiments from Section \ref{sec:metric_verification} are displayed in \cref{fig:groundtruthablation}. We find that our metrics still accurately track the ground-truth sparsity for metric values up to about 32, and can distinguish between the sparse linear and other datasets.

\begin{figure}[hbt]
    \centering
    \includegraphics[width=0.55\textwidth]{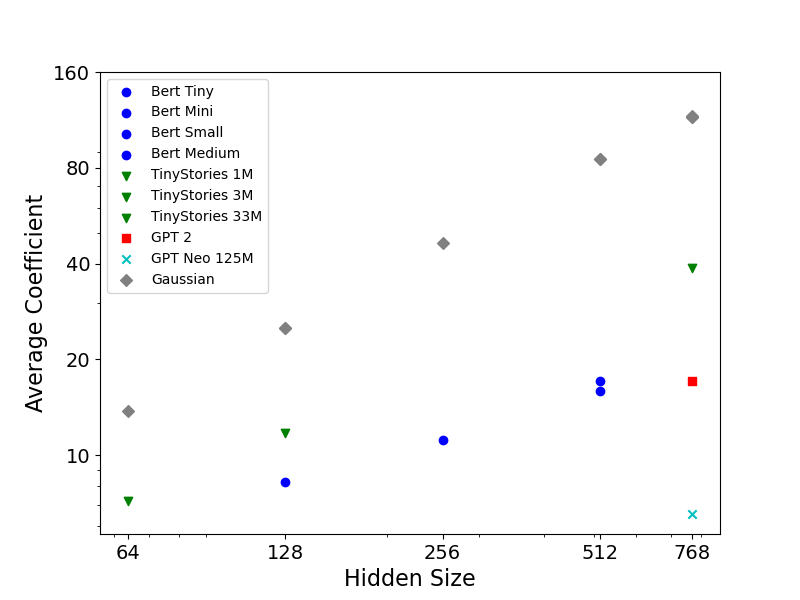}
    \caption{Embedding size versus average coefficient for token embeddings of three different classes of language models.}
    \label{fig:embdding-sparse-averagecoefficient}
\end{figure}

\FloatBarrier

\section{Further Details on Language Model Embedding Experiments}
\label{app:llmsfurtherdetails}

For the experiments in Section \ref{sec:llms} we use a dictionary size of $8d$. We also center the embedding vectors before applying our sparse coding algorithm. After applying our sparse coding algorithm, the decomposition that we find is able to explain over $90\%$ of the variance in the activations across all models.

In \cref{fig:embdding-sparse-averagecoefficient} we show the results of the experiment of Section \ref{sec:embeddinglayers} but using the average coefficient metric rather than normalized loss. We find very similar results to those obtained with normalized loss.

\section{Examples of Interpretable Features in Embedding Space}
\label{app:maxactivatingembedding}

We provide some initial evidence that the feature directions we find via sparse coding do correspond to human-interpretable features, corroborating the findings of \citet{yun2021transformer, cunningham2023sparse}. Given the feature vectors we found in our sparse decomposition of the embeddings of GPT-2, we pick a couple of tokens with clear meanings -- \texttt{season} and \texttt{physics} -- decompose them into their constituent features, pick the top three features for each with the maximum feature coefficient, and then plot the 20 tokens that maximally activate that feature.

\begin{table}[b!]
    \centering
    \def\arraystretch{1.7}
    \begin{tabular}{|c|p{9cm}|c|}
        \hline 
        Feature & \centering{Max Activating examples} & Interpretation \\
        \hline
        Feature 1 & \texttt{episode}, \texttt{Episode}, \texttt{episode}, \texttt{Episode}, \texttt{episodes}, \texttt{Season}, \texttt{isode}, \texttt{isodes}, \texttt{Season}, \texttt{podcast}, \texttt{odcast}, \texttt{podcast}, \texttt{Podcast}, \texttt{season}, \texttt{flashbacks}, \texttt{Seasons}, \texttt{Ep}, \texttt{Ep}, \texttt{epis}, \texttt{season} & TV/Radio shows \\
        Feature 2 & \texttt{Winter}, \texttt{winter}, \texttt{Summer}, \texttt{summer}, \texttt{winter}, \texttt{Winter}, \texttt{Summer}, \texttt{autumn}, \texttt{Autumn}, \texttt{Spring}, \texttt{Spring}, \texttt{summers}, \texttt{winters}, \texttt{spring}, \texttt{Fall}, \texttt{spring}, \texttt{Fall}, \texttt{offseason}, \texttt{seasonal}, \texttt{Halloween} & Yearly seasons \\
        Feature 3 & \texttt{year}, \texttt{Year}, \texttt{YEAR}, \texttt{year}, \texttt{Year}, \texttt{decade}, \texttt{month}, \texttt{years}, \texttt{Years}, \texttt{Month}, \texttt{Years}, \texttt{month}, \texttt{years}, \texttt{yr}, \texttt{Month}, \texttt{Months}, \texttt{months}, \texttt{week}, \texttt{season}, \texttt{months} & Lengths of time \\
        \hline
    \end{tabular}
    \vspace{1mm}
    \caption{Max activating examples for top 3 features in the sparse decomposition of \texttt{season}.}
    \label{tab:pekingdecomposition}
\end{table}

\begin{table}[t!]
    \centering
    \def\arraystretch{1.7}
    \begin{tabular}{|c|p{9cm}|c|}
        \hline 
        Feature & \centering{Max Activating examples} & Interpretation \\
        \hline
        Feature 1 & \texttt{Physics}, \texttt{physics}, \texttt{ysics}, \texttt{physic}, \texttt{Math}, \texttt{Math}, \texttt{Chem}, \texttt{math}, \texttt{Chem}, \texttt{particle}, \texttt{chem}, \texttt{chemistry}, \texttt{asm}, \texttt{physicists}, \texttt{math}, \texttt{particles}, \texttt{asms}, \texttt{Chemistry}, \texttt{maths}, \texttt{Phys} & Physical sciences \\
        Feature 2 & \texttt{biology}, \texttt{psychology}, \texttt{economics}, \texttt{anthropology}, \texttt{neuroscience}, \texttt{sociology}, \texttt{physiology}, \texttt{biology}, \texttt{astronomy}, \texttt{Biology}, \texttt{chemistry}, \texttt{theology}, \texttt{physics}, \texttt{iology}, \texttt{Economics}, \texttt{Chemistry}, \texttt{Anthropology}, \texttt{ecology}, \texttt{mathematics}, \texttt{ochemistry} & Fields of study \\
        Feature 3 & \texttt{interstellar}, \texttt{galactic}, \texttt{galaxies}, \texttt{Interstellar}, \texttt{galaxy}, \texttt{asteroid}, \texttt{Centauri}, \texttt{asteroids}, \texttt{Galactic}, \texttt{astroph}, \texttt{Planetary}, \texttt{Astron}, \texttt{planetary}, \texttt{astronomers}, \texttt{planets}, \texttt{cosmic}, \texttt{spaceship}, \texttt{stellar}, \texttt{Nebula}, \texttt{astronomer} & Astronomy \\
        \hline
    \end{tabular}
    \vspace{1mm}
    \caption{Max activating examples for top 3 features in the sparse decomposition of \texttt{physics}.}
    \label{tab:physicsdecomposition}
\end{table}

\begin{table}[t!]
    \centering
    \def\arraystretch{1.7}
    \begin{tabular}{|c|p{12cm}|}
        \hline
        Token & \hspace{4.2cm} Closest embeddings \\
        \hline
        \texttt{season} & \texttt{season}, \texttt{Season}, \texttt{season}, \texttt{Season}, \texttt{seasons}, \texttt{offseason}, \texttt{preseason}, \texttt{summer}, \texttt{episode}, \texttt{postseason}, \texttt{episodes}, \texttt{Episode}, \texttt{winter}, \texttt{autumn}, \texttt{playoff}, \texttt{Summer}, \texttt{seasonal}, \texttt{podcast}, \texttt{Winter}, \texttt{Ý}, \texttt{Year}, \texttt{subur}, \texttt{Nitrome}, \texttt{StreamerBot}, \texttt{ÃĥÃĤÃĥÃĤÃĥÃĤÃĥÃĤÃĥÃĤÃĥÃĤÃĥÃĤÃĥÃĤÃĥ...}, \texttt{ÃĥÃĤÃĥÃĤÃĥÃĤÃĥÃĤ}, \texttt{externalTo}, \texttt{decade}, \texttt{episode}, \texttt{Seasons} \\
        \hline
    \end{tabular}
    \vspace{1mm}
    \caption{Closest 30 words to \texttt{season} in embedding space.}
    \label{tab:controlmaxactivating}
\end{table}

The results are shown in \cref{tab:pekingdecomposition} and \cref{tab:physicsdecomposition}. We see that for each feature in the decomposition, the maximally activating examples suggest a natural interpretation of that feature, as listed in the third column. For comparison, we also display the 30 words in embedding space which are closest to the embedding direction of \texttt{season} in \cref{tab:controlmaxactivating}. We note that this direction is much more polysemantic than the rows in \cref{tab:pekingdecomposition}, suggesting that our sparse coding method is succeeding at disentangling superposed meanings.

\section{Further Details of Experiments on Later Layers in Language Models}
\label{app:llmlaterlayers}

To form our activation distribution, we sampled model inputs from a dataset of $800$k sentences from Wikipedia abstracts. For each model and layer, we use a dictionary size of $16d$. We fix $\lambda = 0.1$ during training, and then for decomposing into sparse features we use $\lambda=0.0027, 0.0037, 0.0047, 0.01$ for BERT Tiny, Mini, Small and Medium respectively. This allows us to ensure that we explain at least $98\%$ of the variance for all models and layers, while avoiding needing an adaptive choice of $\lambda$ (for which we did not have sufficient computational resources).

\begin{figure}[b!]
    \centering
    \begin{subfigure}[b]{0.48\textwidth}
        \centering
        \includegraphics[width=\textwidth]{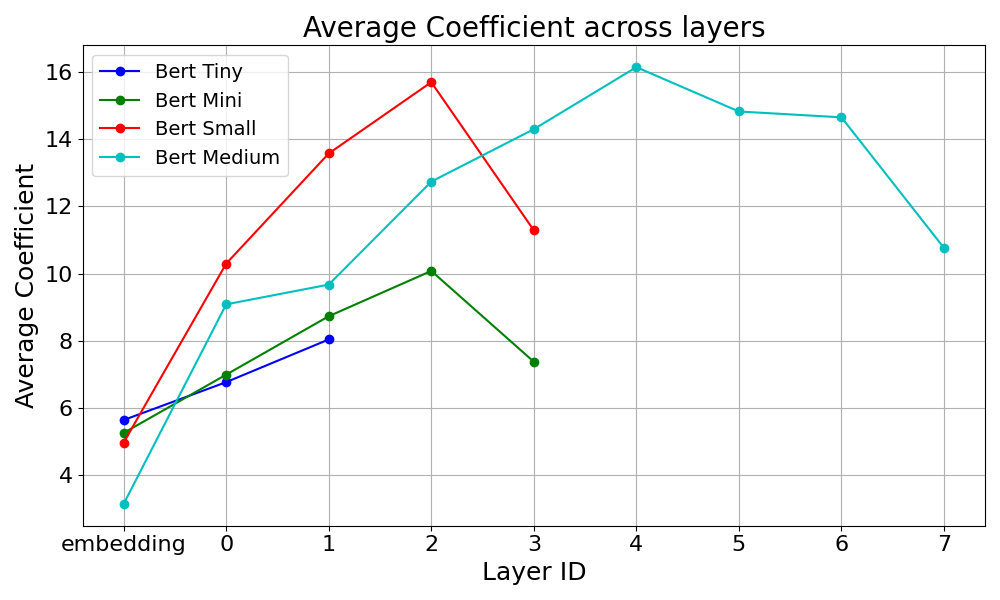}
    \end{subfigure}
    \caption{Sparsity of activations by layer for four BERT models, measured using average coefficient norm.}
    \label{fig:sparsitybylayeraveragecoefficient}
\end{figure}

In \cref{fig:sparsitybylayeraveragecoefficient}, we show the results of the experiment of Section \ref{sec:laterlayers} but using the average coefficient metric rather than the normalized loss. The results are qualitatively similar.

\end{document}